\documentclass[conference]{IEEEtran}
\IEEEoverridecommandlockouts
\usepackage{cite}
\usepackage{amsmath,amssymb,amsfonts}
\usepackage{algorithmic}
\usepackage{graphicx}
\usepackage{textcomp}
\usepackage{xcolor}
\def\BibTeX{{\rm B\kern-.05em{\sc i\kern-.025em b}\kern-.08em
    T\kern-.1667em\lower.7ex\hbox{E}\kern-.125emX}}
\begin{document}

\title{Tiny Eats: Eating Detection on a Microcontroller\\
\thanks{Supported by National Science Foundation under award numbers CNS-1565269 and CNS-1565268. Special thanks to Shengjie Bi and Professor David Kotz at Dartmouth College for their support on the Auracle Project.}
}

\author{\IEEEauthorblockN{Maria T. Nyamukuru}
\IEEEauthorblockA{\textit{Thayer School of Engineering} \\
\textit{Dartmouth College}\\
Hanover, NH, USA \\
Maria.T.Nyamukuru.th@dartmouth.edu}
\and
\IEEEauthorblockN{Kofi M. Odame}
\IEEEauthorblockA{\textit{Thayer School of Engineering} \\
\textit{Dartmouth College}\\
Hanover, NH, USA \\
Kofi.M.Odame@dartmouth.edu}

}

\maketitle

\begin{abstract}
There is a growing interest in low power highly efficient wearable devices for automatic dietary monitoring (ADM) \cite{Prioleau17}. The success of deep neural networks in audio event classification problems makes them ideal for this task. Deep neural networks are, however, not only computationally intensive and energy inefficient but also require a large amount of memory. To address these challenges, we propose a shallow gated recurrent unit (GRU) architecture suitable for resource-constrained applications. This paper describes the implementation of the Tiny Eats GRU, a shallow GRU neural network, on a low power micro-controller, Arm Cortex M0+, to classify eating episodes. Tiny Eats GRU is a hybrid of the traditional GRU \cite{Cho14} and eGRU \cite{Amoh19} which makes it small and fast enough to fit on the Arm Cortex M0+ with comparable accuracy to the traditional GRU. 
The Tiny Eats GRU utilizes only 4\% of the Arm Cortex M0+ memory and identifies eating or non-eating episodes with 6 ms latency and accuracy of 95.15\%.
\end{abstract}


\section{Introduction}
Poor eating habits have been linked to an increased risk of cardiovascular disease, cancer, obesity, diabetes, cataracts, and possibly dementia \cite{Rosenberg09}. 
As such, there is growing interest in recording the food intake of at-risk individuals and exploring diet interventions to prevent long-term health problems \cite{Rosenberg09}. Today's gold standard for recording food intake is the self-administered food diary. This approach depends on an individual's meticulous attention to their daily diet. As such, it is prone to human error and lacks an objective validation check for the self-reported data \cite{Burke11}.

To address these problems, several automated alternatives to the food diary have been proposed. For instance, radio-frequency identification (RFID) equipped devices like the Dietary Intake Monitoring System (DIMS) were introduced in \cite{Ofei14}, but they cannot differentiate between users. Video-based systems are user-specific, but they are bulky, intrusive and require extensive memory and computational resources \cite{Wen09, Cadavid2012}. Wearable devices \cite{Zheng14, Lara13} are another promising approach, but the current wearable implementations have limited computational power, which requires them to transmit raw data for offline processing wirelessly; this continuous wireless streaming severely limits the wearable devices' battery life. In this paper, we introduce a power-efficient shallow GRU neural network for detecting eating episodes that can be implemented directly on a wearable device. Our eating detection model requires no offline processing or wireless communication, thus dramatically improving battery life and helping to make wearable devices a more effective alternative to the food diary.

\section{Eating Detection Overview}
Our system for detecting eating episodes comprises a contact microphone that senses jaw movement \cite{Shengjie18} and a micro-controller that analyzes the microphone output for chewing sounds. The algorithm on the micro-controller is based on the gated recurrent unit (GRU) neural network \cite{Cho14}, which has been applied successfully to various audio event detection problems \cite{Kim17, kusupati2018fastgrnn}. For our algorithm, we modified the GRU activation functions, quantized the weights, and implemented all computations as integer operations. These modifications allow our model to fit within the memory, computational, and latency constraints of low-power embedded processors.

We start by providing details on the neural network architecture in Section \ref{sec:Details}. Next, Section \ref{sec:Results} shows the results of the model implementation and then a discussion in Section \ref{sec:Disc} followed by future work in Section \ref{sec:Fush}. We then summarize the paper in Section \ref{sec:Conc}.

\section{Embedded Neural Network Details}
\label{sec:Details}
\subsection{Arm Cortex M0+}
We implemented our algorithm on the Arm Cortex M0+ because it is the most energy-efficient Arm processor available for constrained embedded applications \cite{Arm}, which makes it ideal for use in small-sized power conservative wearable devices. The challenge of using such a low-power processor is that it has no dedicated floating-point unit, operates at 48 MHz speed, and it has only 32 KB RAM and 256 KB flash memory.  
These limitations restrict the size of the neural network that can be implemented and the size of data that can be computed at a given time.

\subsection{GRU Modifications for Arm Cortex M0+}
\label{subsec:GRU}
In order to meet the resource constraints of the Arm Cortex M0+, we modify the traditional GRU cell \cite{Cho14} and design the Tiny Eats GRU described by:  

\begin{equation}
\begin{aligned}
   \ \textbf{r}_t &= \frac{\varsigma(\textbf{W}_r[\textbf{x}, \textbf{h}_{t-1}]) + 1}{2}\
\end{aligned}
\label{resetgate}
\end{equation}
\begin{equation}
\begin{aligned}
  \  \textbf{z}_t &= \frac{\varsigma(\textbf{W}_z[\textbf{x}, \textbf{h}_{t-1}]) + 1}{2}\
\end{aligned}
\label{updategate}
\end{equation}
\begin{equation}
\begin{aligned}
  \  \Tilde{\textbf{h}}_t &= \varsigma(\textbf{W}_h[\textbf{x}, (\textbf{r}\odot \textbf{h}_{t-1})]) \ 
\end{aligned}
\label{cellstate}
\end{equation}
\begin{equation}
\begin{aligned}
  \  \textbf{h}_t &= \textbf{z}_t\textbf{h}_{t-1} + (1 -  \textbf{z}_t)\Tilde{\textbf{h}}_t, \
\end{aligned}
\label{candidatestate}
\end{equation}
where \textbf{x} represents the input features, \textbf{W} represents the 8 bit quantized weights, \textbf{r} represents the reset gate, \textbf{z} represents the update gate, \begin{math}\Tilde{\textbf{h}}\end{math} represents the candidate state, \textbf{h} represents the new cell state, and \begin{math}\odot\end{math} represents the element-wise multiplication of two vectors.

The Tiny Eats GRU cell diverges from the traditional GRU cell in that it uses a shifted soft-sign (\begin{math}\varsigma\end{math}) activation function in place of sigmoid activation function for the update and reset gates. A regular soft-sign is also used in place of tanh activation functions for candidate state as proposed by eGRU cell \cite{Amoh19}. The sigmoid and tanh activation functions are memory- and resource-intensive since they require iterative accesses of look-up tables. The soft-sign is a more computationally-conservative alternative compared to the sigmoid and tanh. This computation is particularly crucial because the Arm Cortex M0+ lacks a floating-point unit (FPU) and Digital Signal Processing (DSP) instruction set. 
The Tiny Eats GRU cell diverges from the eGRU cell \cite{Amoh19} in that it maintains the use of a reset gate as proposed by the traditional GRU cell. This helps extract the portion of the history that is relevant to the current computation, as highlighted in equation \eqref{cellstate}.

\subsection{Embedded Neural Network}
\begin{figure}[htbp]
\hspace*{1.0cm}
  \includegraphics[width=7cm, height=7cm]{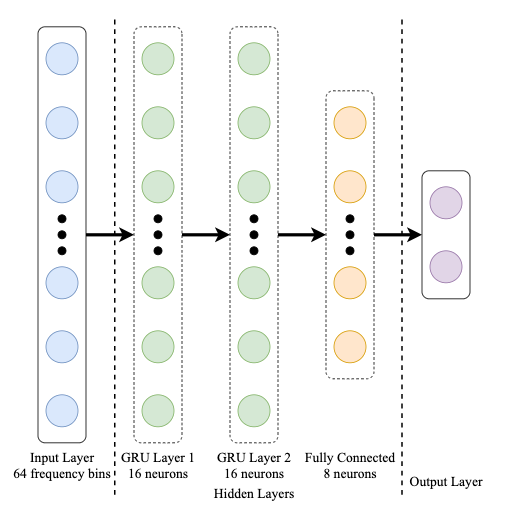}
  \caption{Gated Recurrent Unit Neural Network Architecture}
  \label{fig:NN}
\end{figure}

The universal approximation theorem states that simple neural networks can represent a wide variety of interesting functions when given appropriate parameters. Instead of using a standard deep neural network, we use a shallow neural network with fewer layers and fewer neurons to represent the differences in STFT parameters of {\itshape eating} and {\itshape non-eating} data. The shallow neural network is computationally cheaper and conserves memory, which is necessary for implementation on the Arm Cortex M0+. In this paper, we propose the shallow neural network architecture shown in Fig. \ref{fig:NN} that consists of 1 output layer and only 3 hidden layers; 2 gated recurrent units designed as explained in Section \ref{subsec:GRU}, each with 16 neurons, and one fully connected layer with 8 neurons. The fully connected layer consists of a linear pre-activation function and a soft-sign activation. The output layer consists of a linear pre-activation function and a soft-max activation function. This design is very similar to the eGRU suggested by {\itshape Amoh and Odame} \cite{Amoh19}.

\section{Experiments and Results}
\label{sec:Results}
\subsection{Eating Detection Dataset}
\label{subsec:STFT}

The data used for this study was previously collected in a controlled laboratory setting from 20 participants using a contact microphone placed behind the participant's ear \cite{Shengjie17}. The contact microphone was connected to a data acquisition device (DAQ) that collected the activity data in real-time at a 20KHz sampling rate with 24-bit resolution. The contact microphone, used for acoustic sensing, is placed at the tip of the mastoid bone and secured using a headband to ensure contact with the body. The participants were instructed to perform an activity that involves both {\itshape eating} and {\itshape non-eating}. {\itshape Eating} activities included eating carrots, protein bars, crackers, canned fruit, instant food, and yogurt, sequentially for 2 minutes per food type. This resulted in a 4 hour total {\itshape eating} dataset. 
{\itshape Non-eating} activities included talking and silence for 5 minutes each and then coughing, laughing, drinking water, sniffling, and deep breathing for 24 seconds each. This resulted in a 7 hour total {\itshape non-eating} dataset. Each activity occurred separately and was classified based on activity type as {\itshape eating} or {\itshape non-eating}.

\begin{figure}[htbp]
  \includegraphics[width=8.5cm, height=5.5cm]{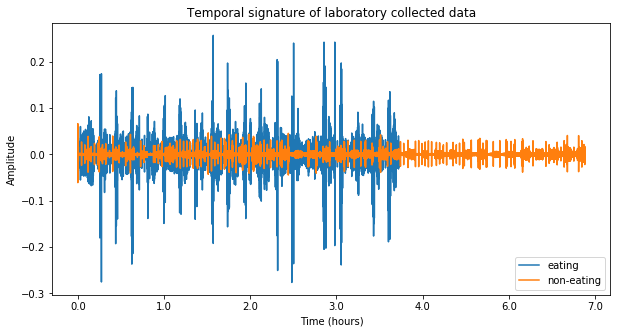}
  \caption{Temporal signature of down-sampled and filtered laboratory collected raw data}
\label{fig:filtered raw}
\end{figure}

\subsection{Data Pre-Processing}
\label{subsec:Data}
\begin{figure}[htbp]
  \includegraphics[width=8.5cm, height=5cm]{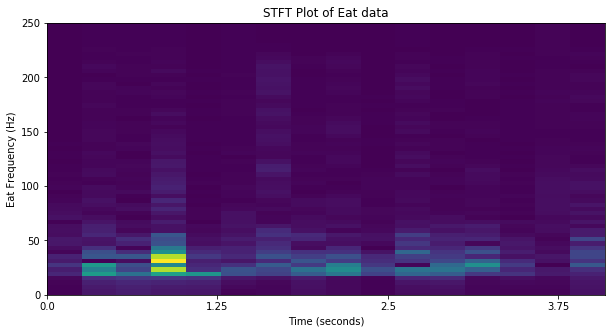}
  \caption{STFT Results of Eating data}
  \label{fig:STFTa}
\end{figure}

\begin{figure}[htbp]
  \includegraphics[width=8.5cm, height=5cm]{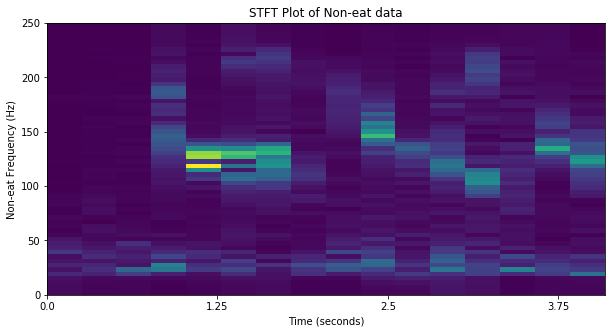}
  \caption{STFT Results of Non-Eating data}
  \label{fig:STFTb}
\end{figure}

In this paper, we down-sampled the raw sound data to 500 Hz to reduce power consumption and memory usage and applied a high pass filter with a 20Hz cutoff frequency to attenuate noise \cite{Shengjie17} as shown in Fig. \ref{fig:filtered raw}. We segmented the positive class data ({\itshape eating data}), and negative class data ({\itshape non-eating data}) into 4-second windows, and computed a Short Time Fourier Transform (STFT), with 128 Fast Fourier Transform bins and no overlap, over the segments to view and extract features as shown in Fig. \ref{fig:STFTa} and Fig. \ref{fig:STFTb}.

The positive class data, as shown in Fig. \ref{fig:STFTa} has a lower frequency and periodic envelope distribution in comparison to the negative class data shown in Fig. \ref{fig:STFTb} that has a higher frequency and more compact distributions. These STFT features were used to train the neural network with each 4-second window being classified as an {\itshape eating} episode or a {\itshape non-eating} episode.

\subsection{Neural Network Training}
\label{subsec:NNTraining}
The neural network shown in Fig. \ref{fig:NN} is first programmed and trained in Python using TensorFlow Keras v2.0. This model uses floating-point. The training dataset contains more representation of {\itshape non-eating} data than it does {\itshape eating} data as described in Section \ref{subsec:STFT}. This imbalance inherently creates a bias in the training algorithm. In order to debias the algorithm, we normalize the binary cross-entropy loss function with each class's representative weight during training. This ensures that both classes are represented fairly.

\begin{figure}[htbp]
  \includegraphics[width=8.8cm, height=5.5cm]{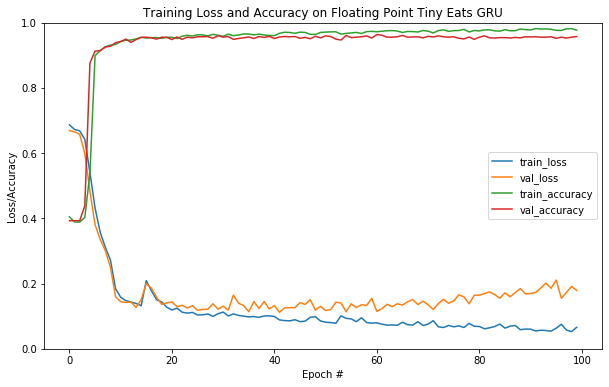}
  \caption{Loss and Accuracy Training Graphs for Floating Point Tiny Eats GRU}
  \label{fig:fpu_training}
\end{figure}

In order to guarantee that the best performing model is used, a model checkpoint is implemented to save and retrieve the model with the highest validation accuracy and minimal validation loss. The results of this FPU model are as shown in Table \ref{tab:Acc} and Fig. \ref{fig:fpu_training}. 

\begin{figure}[htbp]
  \includegraphics[width=8.8cm,height=5.5cm]{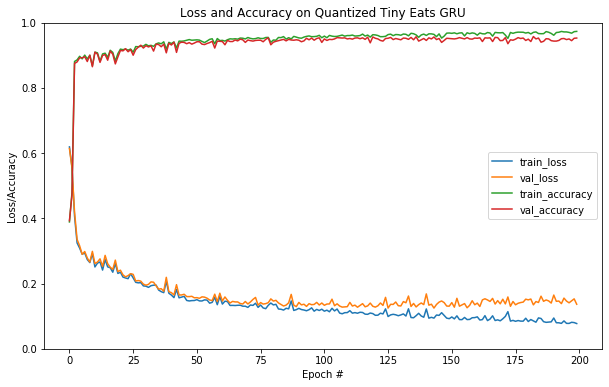}
  \caption{Quantized (8 bit) Tiny Eats GRU Loss and Accuracy Training Graphs}
  \label{fig:quantized_accuracy}
\end{figure}

An integer quantization based model \cite{Amoh19} that uses 8-bit quantization for all weights is then implemented in Pytorch and trained, as shown in Fig. \ref{fig:quantized_accuracy}. The weight normalization applied to the cross-entropy loss function, and the model checkpointing used in the FPU model is implemented here as well.

The cross-validation accuracy and loss from both models are compared, as shown in Table \ref{tab:Acc}.

\begin{table}[htbp]
\caption{Comparison of FPU and Quantized Tiny Eats GRU Cross Validation Accuracy}
\begin{center}
\begin{tabular}{|c|c|c|c|}
\hline
\textbf{Model}&\multicolumn{2}{|c|}{\textbf{Model Designs}} \\
\cline{2-3} 
\textbf{Metrics} & \textbf{\textit{FPU}}& \textbf{\textit{Quantized}} \\
\hline
Cross-Validation Accuracy & 96.13\% & 94.41\% \\
Cross-Validation Loss & 0.12 & 0.13 \\ 
Epochs Used & 100 & 200 \\
\hline
\end{tabular}
\label{tab:Acc}
\end{center}
\end{table}

\begin{table}[htbp]
\caption{Evaluation of Tiny Eats GRU on independent Test set vs other models}
\begin{center}
\begin{tabular}{|c|c|c|c|c|c|c|c|}
\hline
\textbf{Model}&\multicolumn{5}{|c|}{\textbf{Models Used}} \\
\cline{2-6} 
\textbf{Measure} & \textbf{\textit{Tiny Eats}}& \textbf{\textit{Model in }} & \textbf{\textit{Ear Bit}}& \textbf{\textit{Splendid}}& \textbf{\textit{EGG}}\\
 & & \cite{Shengjie17} & \cite{Bedri2017} & \cite{Papapanagiotou2017} & \cite{Farooq2014} \\
\hline
    Model & GRU & LR & RF & SVM & FFNN  \\
    Test Acc. & 95.15\% & 91.5\% & 90.1\% & 93.8\% & 90.1\% \\ 
    F1-Score & 94.89\% & NA & 90.9\% & 76.1\% & NA \\ 
    Precision & 94.68\% & 95.1\% & 86.2\% & 79.4\% & 88.4\% \\ 
    Recall & 95.12\% & 87.4\% & 96.1\% & 80.7\% & 91.8\% \\ 
\hline
\end{tabular}
\label{tab:Comparisons}
\end{center}
\end{table}

The quantized model is evaluated on an independent test set, and the test accuracy, loss, precision, recall, and F1-score are reported, as shown in Table \ref{tab:Comparisons}. The precision score represents the ratio of correctly classified positive observations of the total number of observations classified as positive. The recall score (sensitivity) represents the ratio of correctly classified positive observations to the total number of true positive observations, as shown in equation \eqref{precision} and equation \eqref{recall} where $TP$ is True Positive, $FN$ is false negative, and $FP$ is false positive.

\begin{equation}
\begin{aligned}
   \ Precision &= \frac{TP}{TP + FP}\
\end{aligned}
\label{precision}
\end{equation}
\begin{equation}
\begin{aligned}
  \ Recall &= \frac{TP}{TP + FN}\
\end{aligned}
\label{recall}
\end{equation}

The trained integer quantized model is then implemented on the Arm Cortex M0+.
 
\subsection{Arm Cortex M0+ Implementation}
\label{subsec:ArmImp}
We perform the feature extraction on the Arm Cortex M0+ by implementing a Fast Fourier Transform which utilizes only 9\% (23 KB) of the memory.
The features generated from the Arm FFT are then transmitted to the implemented Tiny Eats GRU. The implementation of the GRU on the Arm Cortex M0+ takes 6 ms to execute one sample and utilizes only 4\% (12 KB) of the memory.

\begin{figure}[htbp]
  \includegraphics[width=8.8cm,height=5.2cm]{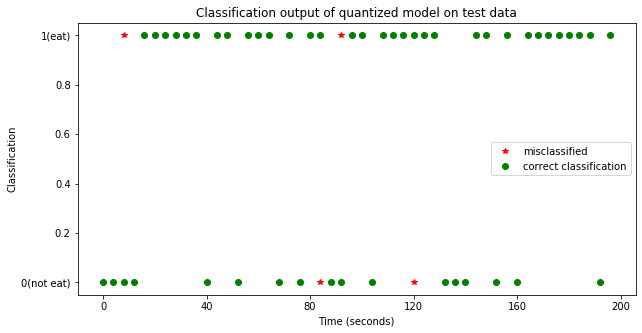}
  \caption{Quantized (8bit) Tiny Eats GRU Predicted label vs true label}
  \label{fig:quantized_output}
\end{figure}

The accuracy of the GRU implementation is verified by transmitting test data via serial communication and reading the output label classification. This was done with a subset of the test data set, which recorded a label match between the Pytorch quantized model and the Arm Cortex M0+ implementation, as shown in Fig. \ref{fig:quantized_output}.

\section{Discussion}
\label{sec:Disc}
In order to account for the limitation in the size of the neural network used, advanced feature engineering is paramount. The Tiny Eats GRU utilizes STFT features shown in Section \ref{subsec:STFT} as the only class of input features. This is crucial because it can be implemented on the Arm Cortex M0+ at a low cost, as shown in Section \ref{subsec:ArmImp}, while ensuring significant accuracy is achieved. The efficient implementation of the STFT on the Arm Cortex M0+ reduces the burden on the neural network. It consequentially allows for a reduction in the number of layers and the number of neurons required for classification. This minimization of neural network parameters is resource-efficient and allows for implementation on resource-constrained micro-controllers while achieving high accuracy.

In order to reduce memory usage and run time execution performance, we utilize quantized weights instead of floating-point weights and fixed-point arithmetic. Using 8-bit quantized weights reduces the memory footprint of the network, and when coupled with fixed-point arithmetic, reduce the power consumption and increases computational speed, with minimal impact on the accuracy of the overall network, making it efficient for deployment on embedded hardware \cite{Han2016, Vanhoucke2011}. As shown in Table \ref{tab:Acc}, there is only a 2\% decrease in accuracy by switching to 8-bit integer quantization from floating-point. 8-bit weight quantizations also ensure that the Tiny Eats GRU is small enough to fit on the Arm Cortex M0+ and run efficiently, as seen in Section \ref{subsec:ArmImp}.

In order to increase computational speed, the Tiny Eats GRU uses soft-sign activations functions, as explained in Section \ref{subsec:GRU}. The results shown in Section \ref{subsec:NNTraining} affirm that the proposed Tiny Eats GRU architecture is a viable replacement for the traditional GRU \cite{Cho14} in eating episode classification. The GRU network with fast soft-sign activation functions achieves a cross-validation accuracy of 96.13\% and 94.41\% for floating-point and quantized weight implementation, respectively, as shown in Table \ref{tab:Acc}. This soft-sign implementation achieves high accuracy while minimizing computation time.

The proposed Tiny Eats GRU's performance outperforms some of the recently proposed approaches to eating detection, as shown in Table \ref{tab:Comparisons}. The accuracy, precision, recall, and F1-score of the Ear Bit Sensor \cite{Bedri2017} that uses a random forest (RF) model,  the SPLENDID device \cite{Papapanagiotou2017} that uses support vector machine (SVM) classification, and the electroglottography (EGG) device \cite{Farooq2014}  that utilizes feedforward neural networks (FFNN), are compared to that of the Tiny Eats GRU. The Tiny Eats GRU has better overall performance on all evaluation metrics, recording the highest F1-score of 94.89\%, as shown in Table \ref{tab:Comparisons}. These results are comparable to the model design in \cite{Shengjie17} that uses linear regression (LR).

The Tiny Eats GRU is extremely useful in resource-constrained environments like the Arm Cortex M0+, which can be used in low power wearable devices. It operates with only 6 ms latency, which is significantly smaller than the 250 ms FFT feature extraction time and uses only 4\% of the M0+ memory, which allows for the execution of other processes that could include implementing more features.

\section{Future Work}
\label{sec:Fush}
In the future, we can further improve the accuracy by increasing the number of feature classes to represent more relationships besides the STFT. Feature classes like the zero-crossing rate and number of peaks have defining characteristics that are essential in classifying {\itshape eating} and {\itshape non-eating} data. This would also imply that we could reduce the number of neurons required in the Tiny Eats GRU, and consequentially reducing its computation time and memory requirements, making it even more efficient.

 Furthermore, we will implement and train this algorithm on data collected outside the lab to infer its performance in an unconstrained, free-living environment. 
 We will also explore classifying different food types and inferring the volume of food eaten during an eating session to gain more insight into the user's diet.

\section{Conclusion}
\label{sec:Conc}
This paper proposed the Tiny Eats GRU architecture, an implementation on Arm Cortex M0+ for the detection of eating episodes. The Tiny Eats GRU was demonstrated to have significantly low classification costs by being memory-conservative, computationally efficient with significant accuracy results while consuming minimal power. This allows for the implementation of Tiny Eats GRU on severely resource-constrained devices incapable of handling traditional deep neural networks, that can be used in wearable devices.

\section*{Acknowledgment}

This work was supported by the National Science Foundation under award numbers CNS-1565269 and CNS-1565268. We would like to thank Shengjie Bi and Professor David Kotz at Dartmouth College for their support on the Auracle Project. The views and conclusions contained in this document are those of the authors and should not be interpreted as necessarily representing the official policies, either expressed or implied, of the sponsors.

\bibliographystyle{./bibliography/IEEEtran}
\bibliography{./bibliography/TEGRU}

\end{document}